\documentclass[letterpaper, 10 pt, conference]{ieeeconf}  

\IEEEoverridecommandlockouts                              

\usepackage[dvipsnames]{xcolor}

\usepackage{graphicx}
\usepackage{amsmath}
\usepackage{amssymb}
\usepackage{booktabs}

\usepackage{multirow}
\usepackage{algorithm}
\usepackage{algorithmic}
\usepackage[T1]{fontenc}
\usepackage{multicol}
\usepackage{stfloats}
\usepackage{xcolor}

\usepackage{mathtools}

\DeclarePairedDelimiter\floor{\lfloor}{\rfloor}
 
\usepackage{makecell}
\usepackage{scalerel}
\usepackage{array}
\usepackage{dsfont}
\usepackage{url}
\usepackage{pifont}
\usepackage{adjustbox}
\usepackage{hyperref}
\usepackage{cleveref}
\let\labelindent\relax
\usepackage[shortlabels]{enumitem}

\newcolumntype{C}[1]{>{\centering\let\newline\\\arraybackslash\hspace{0pt}}p{#1}}
\newcolumntype{R}[1]{>{\raggedleft\let\newline\\\arraybackslash\hspace{0pt}}p{#1}}
\newcolumntype{L}[1]{>{\raggedright\let\newline\\\arraybackslash\hspace{0pt}}p{#1}}
\newcolumntype{M}[1]{>{\centering\let\newline\\\arraybackslash\hspace{0pt}}m{#1}}
\newcommand*\rot{\rotatebox{90}}
\newcommand\Tstrut{\rule{-3pt}{2.6ex}}       
\newcommand\Bstrut{\rule[-0.9ex]{-3pt}{0pt}} 
\newcommand{\TBstrut}{\rule{-3pt}{2.6ex} \rule[-0.9ex]{-2pt}{0pt}}  

\newcommand\mydots{\hbox to 1em{.\hss.\hss.}}

\definecolor{myblue}{RGB}{0, 250, 0}
\definecolor{mypink}{RGB}{237, 2, 140}
\definecolor{green1}{RGB}{148, 193, 117}
\definecolor{green2}{RGB}{90, 138, 57}
\definecolor{red}{RGB}{200, 20, 0}
\definecolor{blue}{RGB}{0, 20, 200}

\newcolumntype{D}[2]{%
    >{\adjustbox{angle=#1,lap=\width-(#2)}\bgroup}%
    l%
    <{\egroup}%
}

\title{\LARGE \bf
Embedding Pose Graph, Enabling 3D Foundation Model Capabilities with a Compact Representation.
}

\author{
Hugues Thomas,
Mouli Sivapurapu,
Jian Zhang
\thanks{Hugues Thomas and Jian Zhang are with Apple, Cupertino, USA}
}

\begin{document}

\maketitle
\thispagestyle{empty}
\pagestyle{empty}

\begin{abstract}

This paper presents the Embedding Pose Graph (EPG), an innovative method that combines the strengths of foundation models with a simple 3D representation suitable for robotics applications. Addressing the need for efficient spatial understanding in robotics, EPG provides a compact yet powerful approach by attaching foundation model features to the nodes of a pose graph. Unlike traditional methods that rely on bulky data formats like voxel grids or point clouds, EPG is lightweight and scalable. It facilitates a range of robotic tasks, including open-vocabulary querying, disambiguation, image-based querying, language-directed navigation, and re-localization in 3D environments. We showcase the effectiveness of EPG in handling these tasks, demonstrating its capacity to improve how robots interact with and navigate through complex spaces. Through both qualitative and quantitative assessments, we illustrate EPG's strong performance and its ability to outperform existing methods in re-localization. Our work introduces a crucial step forward in enabling robots to efficiently understand and operate within large-scale 3D spaces.

\end{abstract}

\section{INTRODUCTION}

Recent advances in artificial intelligence have seen the rise of foundation models, a class of large-scale machine learning models that are pre-trained on vast amounts of data and can be fine-tuned for a wide range of downstream tasks. Natural language processing has been revolutionized with the GPT architecture \cite{brown2020language}, and computer vision has also seen tremendous advances with open-vocabulary tasks \cite{radford2021learning} and unsupervised learning \cite{caron2021emerging}. Examples like PALM-E \cite{driess2023palm}, Anyloc \cite{keetha2023anyloc}, or LERF \cite{kerr2023lerf, rashid2023language} showcase the profound impact of such models in robotics. However, while large language models have advanced drastically, their spatial understanding remains limited, particularly in the context of three-dimensional representations. The capacity to reason about and navigate within large-scale 3D environments is a critical frontier in robotics, one that poses unique challenges in terms of data representation and computational efficiency. The complexity of large-scale 3D scene understanding arises from the sheer volume of data associated with detailed environment representations and the delicate balance needed between resolution and comprehensive scene coverage. Traditional methods like voxel grids, point clouds, or meshes can be unwieldy and inefficient when it comes to scalability and real-time processing. 

Addressing these challenges, this paper introduces the EPG, an innovative approach that distills foundational model capabilities into a compact 3D representation. At the core of EPG is a simple but powerful idea: attaching foundation model features to nodes in a pose graph. In contrast to many other works that focus on attaching features to 3D locations (points, voxels, meshes), this idea does not require complex reprojection heuristics and remains lightweight. The EPG framework enables a rich encapsulation of the environmental context while maintaining a sparse and manageable data structure, suitable for various robotic applications. The features attached to each node can vary in nature and serve different purposes depending on the downstream tasks. In this work, we highlight the impact of two types of features for semantic understanding and localization, but more downstream tasks can be explored in future research. With EPG, we provide robots with a foundational spatial understanding that can be dynamically scaled in resolution and size according to the task demands.

\begin{figure}[b]
    \vspace{-3ex}
    \centering
    \adjincludegraphics[width=0.99\columnwidth,trim={{.08\width} {.04\width} {.04\width} {.02\width}},clip]{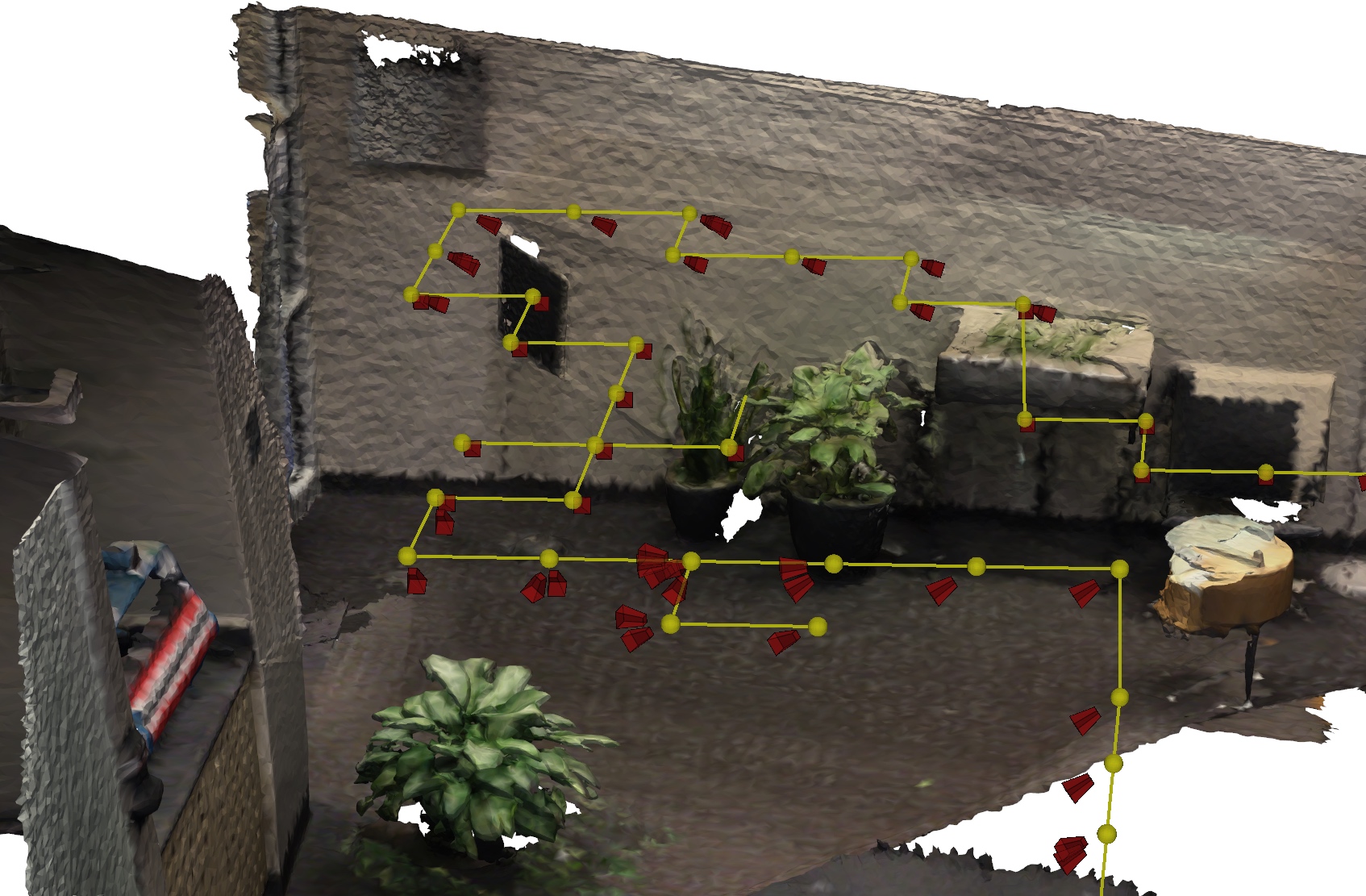}
    \caption{An Embedding Pose Graph visualized with the scene mesh on the ScanNet dataset.}
    \label{fig:example_intro}
\end{figure}

\begin{figure*}[t!]
    \centering
    \adjincludegraphics[width=0.99\textwidth,trim={{.03\width} {.68\height} {.03\width} {.002\height}},clip]{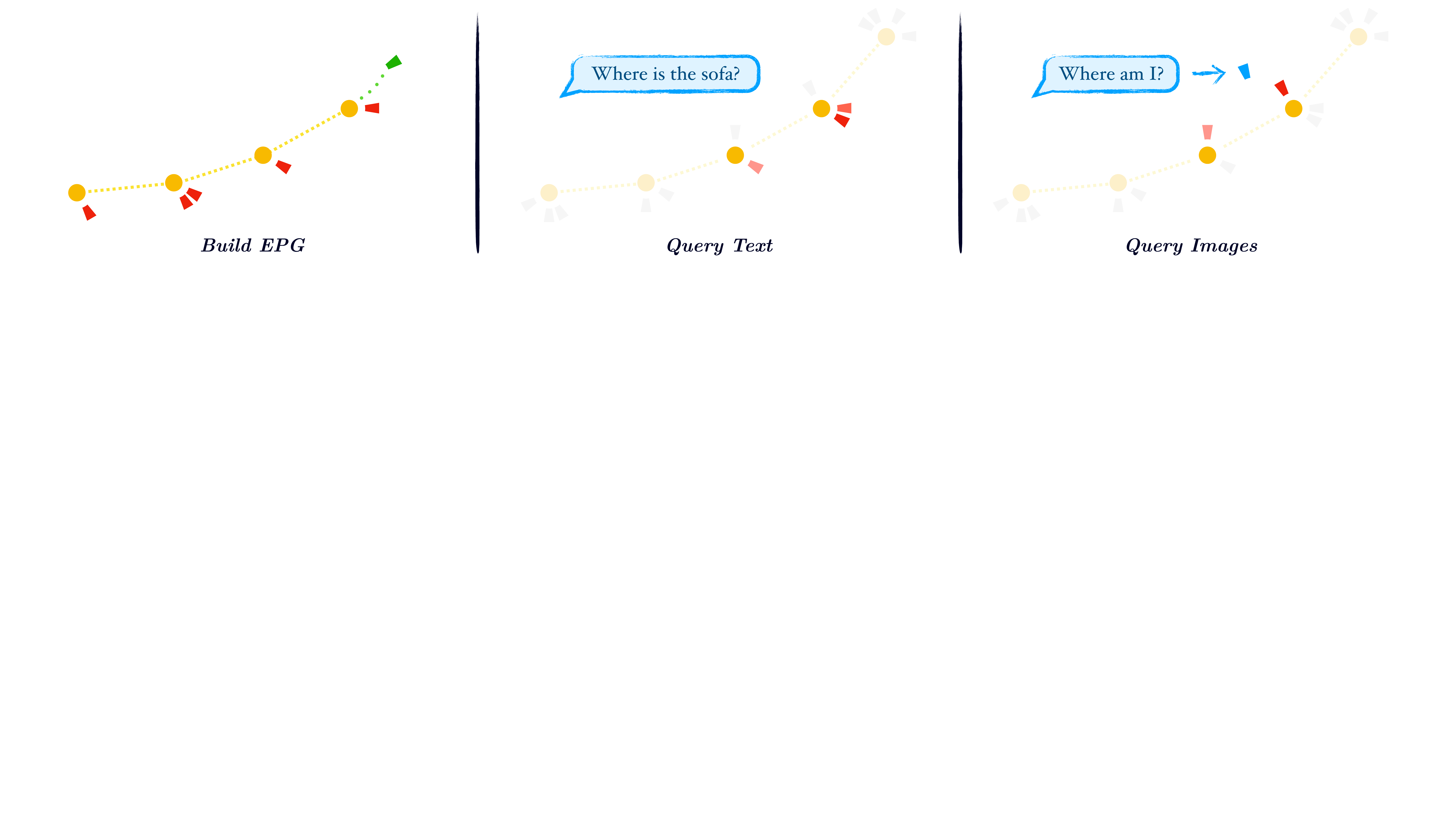}
    \caption{EPG is a compact and versatile tool for spatial understanding in various robotics applications. Once built for a 3D scene, it allows efficient text or image query.}
    \label{fig:intro}
    \vspace{-3ex}
\end{figure*}

In this paper, we first detail the construction of the EPG in \Cref{sec:building}. Although foundation model embeddings could eventually be incorporated into the optimization of the pose graph, aiding both odometry and loop closure, this is not the focus of this paper. Therefore, we assume that we have already collected a set of camera images and their associated poses using a standard simultaneous localization and mapping (SLAM) approach. For our real-world experiments, we use ORB-SLAM2 \cite{mur2015orb, mur2017orb}, and when available in datasets, we use the provided poses for each image. Our building process minimizes redundant data representation through intelligent pose sampling. We subdivide the 3D space into a spatial grid, and within each cell, we further subdivide the camera rotation by yaw ($\theta$) and pitch ($\phi$). In each cell of this 5D grid, a single pose and its corresponding embeddings are stored. We select the pose that is closest to the cell center, to ensure efficient coverage of the 3D space. As foundation model features, we opt to use CLIP for semantic understanding \cite{radford2021learning} and PCA-VLAD-DinoV2 features for localization, similar to \cite{keetha2023anyloc}.

In \Cref{sec:using}, we detail the various tasks that EPG can handle after its construction. EPG demonstrates its versatility across a range of robotic tasks, providing a robust solution for open-vocabulary querying, disambiguation, language navigation, image querying, and re-localization within large-scale 3D environments. Specifically, EPG's ability to facilitate open-vocabulary querying allows robots to interpret and respond to natural language commands by finding relevant poses within the 3D environment that match the queried words or sentences. Furthermore, EPG enhances the robot's ability to disambiguate queries in cluttered or densely populated scenes, effectively managing situations where multiple instances of the requested object are present. In terms of navigation, EPG significantly simplifies the process by guiding robots to previously captured poses relevant to the task at hand, thereby avoiding complex positioning heuristics. Additionally, the image querying capability introduces a powerful tool for spatial recognition and localization, enabling robots to identify their position relative to an input image, even under significant viewpoint changes. Collectively, these applications of EPG underscore its potential to transform how robots interact with and navigate through complex 3D spaces.

Finally, we present qualitative and quantitative results for these tasks in \Cref{sec:exp}. We provide insights into the EPG building process, and how view redundancy is affected by parameter changes. Additionally, we show multiple examples of text queries and disambiguations and analyze the re-localization performances of our approach against state-of-the-art methods in both indoor and outdoor environments. Our approach demonstrates superior re-localization performance on the ScanNet \cite{dai2017scannet} and KITTI \cite{geiger2012we} datasets compared to existing state-of-the-art methods.

Our contributions are as follows:
\begin{itemize}
    \itemsep 0ex 
    \item We introduce the Embedding Pose Graph (EPG) a compact, versatile representation of 3D environments, with foundation model features.
    \item We demonstrate several applications of EPG, showcasing its potential to serve as a foundational block for the spatial understanding required in robotics. These applications include open-vocabulary queries, disambiguation tasks, language-directed navigation, image-based queries, and re-localization capabilities within pre-mapped environments.
\end{itemize}

\section{RELATED WORK}
\label{sec:related}

{\bf Foundation Models for 3D Scenes.}
Several works have been proposed to extract and store foundation model features in a 3D scene. Methods exist that project foundation model features onto 3D point cloud scene representations. VLMaps \cite{huang2023visual} creates a map with visual-language features for open-vocabulary landmark indexing. OpenScene \cite{peng2023openscene} proposes using multiview feature fusion and 3D feature distillation for open-vocabulary queries on 3D point cloud scenes. ConceptFusion \cite{jatavallabhula2023conceptfusion} builds open-set 3D maps that can be queried via text, clicks, images, or audio offline. Structures other than point clouds can also be utilized. Open-Fusion \cite{yamazaki2023open} employs a Truncated Signed Distance Function (TSDF) to define the scene and carry features. CLIP-Fields \cite{shafiullah2022clip} and LERF \cite{kerr2023lerf} train scene-specific neural fields that embed vision-language features. Some works also combine these types of representations with Large Language Models (LLMs) to perform robotics tasks \cite{gadre2023cows, chen2023open}. However, all these methods rely on the concept of projection. The foundation model features are not retained and associated with their original sources, which are images at specific view poses. This can lead to projection errors and often results in very heavy representations of scenes. In contrast, we propose representing the scene through a collection of view poses and their corresponding embeddings, which, in most scenarios, is more lightweight and easier to use.

{\bf Foundation Models for Robotics.}
In this discussion, we explore the use of foundation models and LLMs for general robotics control and refer to~\cite{firoozi2023foundation} for a more in-depth survey. Multiple approaches have been proposed to leverage the strengths of LLMs for robotics tasks. 
Palm-e~\cite{driess2023palm} suggests inputting tokens from various modalities, such as images, neural 3D representations, or states, along with text tokens, into LLMs. The model then generates high-level robotics instructions for tasks including mobile manipulation, motion planning, and tabletop manipulation. In contrast, Instruct2Act~\cite{huang2023instruct2act} generates Python programs that form a complete perception, planning, and action loop for robotic tasks. Moreover, RT-2~\cite{brohan2023rt} generates low-level actions for robots, enabling closed-loop control for visual navigation in unseen environments. Several other studies on visual navigation using LLMs have been conducted. Work~\cite{zhou2023esc,chen20232} considers using text-only LLMs for semantic parsing, followed by feeding the parsed semantics into subsequent Vision-Language Models (VLMs). L3MVN~\cite{yu2023l3mvn} proposes a method that calculates the entropy of objects in each frontier using a semantic segmentation model. This entropy is represented as query strings, and LLMs are used to determine a more relevant frontier. NavGPT~\cite{zhou2023navgpt} and another recent approach~\cite{vemprala2023chatgpt} interact with different visual foundation models to handle multimodal inputs. Closer to our work, LM-Nav~\cite{shah2023lm} utilizes pre-trained foundation models to extract and save embeddings at different locations in the environment for language-based navigation. However, they are limited to semantic features and language queries. On the other hand, AnyLoc~\cite{keetha2023anyloc} proposes using localization features for re-localization but does not include semantic information to allow language interactions. Furthermore, it does not propose an efficient mechanism for constructing the data structure that holds the embeddings. Our approach combines both semantic and localization features and focuses on the efficient construction of the EPG, which allows complete interaction with the scene, including text and image queries.

\section{BUILDING OF AN EPG} 
\label{sec:building}

\subsection{Initial Setup and Data Structure}

In this paper, we intentionally leave out the pose graph optimization problem and consider a simple setup where the EPG is built offline after a navigation session has been captured. Large-scale scenes can be obtained by repeating the process multiple times at different locations. This choice is motivated by the fact that a SLAM pose graph and an EPG serve different purposes. While the SLAM graph needs to keep redundant poses for loop closure edges, an EPG requires sparsity to remain compact as a 3D representation. Separating the two is an effective way to showcase the capabilities of an EPG, and we leave the question of a unified SLAM-EPG representation for future work.

To build an EPG, we start with a dataset composed of $N$ images captured by an RGB or RGB-D camera, each associated with a precise camera pose identified through a $4\times4$ pose matrix. For our optional local refinement method, we assume that the data come with depth images and a 3D point cloud of the scene. 

At the core of our approach is a 5-dimensional grid partitioning the spatial domain along $\left(x, y, z\right)$, and the orientation domains through angular parameters, $\theta$ and $\phi$ (respectively yaw and pitch). This grid, parameterized by $dl$, $d\theta$, and $d\phi$, is designed to host a single pose and its correlated embedding for each segmented cell, ensuring a compact representation of the three-dimensional scene. For the spatial components $x$, $y$, and $z$, we adopt a linear subdivision where the indices $\left(i, j, k\right)$ correlate directly with the position coordinates, scaled by the grid resolution $dl$:

\begin{equation} \label{eq:1}
    \left(i, j, k\right) = \floor*{\frac{\left(x, y, z\right)}{dl}}  \: .
\end{equation}

The angular domains, however, necessitate a more nuanced partitioning scheme to maintain cell uniformity across the sphere's surface. We achieve this through a spherical partitioning method where each $\phi$ segment forms a ring of varying circumference around the sphere, and for each ring, the number of $\theta$ divisions is proportional to its circumference. For the top and bottom rings, we do not subdivide them. This ensures relatively consistent cell size across different latitudes of the spherical surface, with single cells capping the sphere at its poles, as shown in \Cref{fig:grid}. The complete partitioning rule is:

\begin{figure}[b]
    \centering
    \vspace{-3ex}
    \adjincludegraphics[width=0.99\columnwidth,trim={{.001\width} {.001\height} {.001\width} {.001\height}},clip]{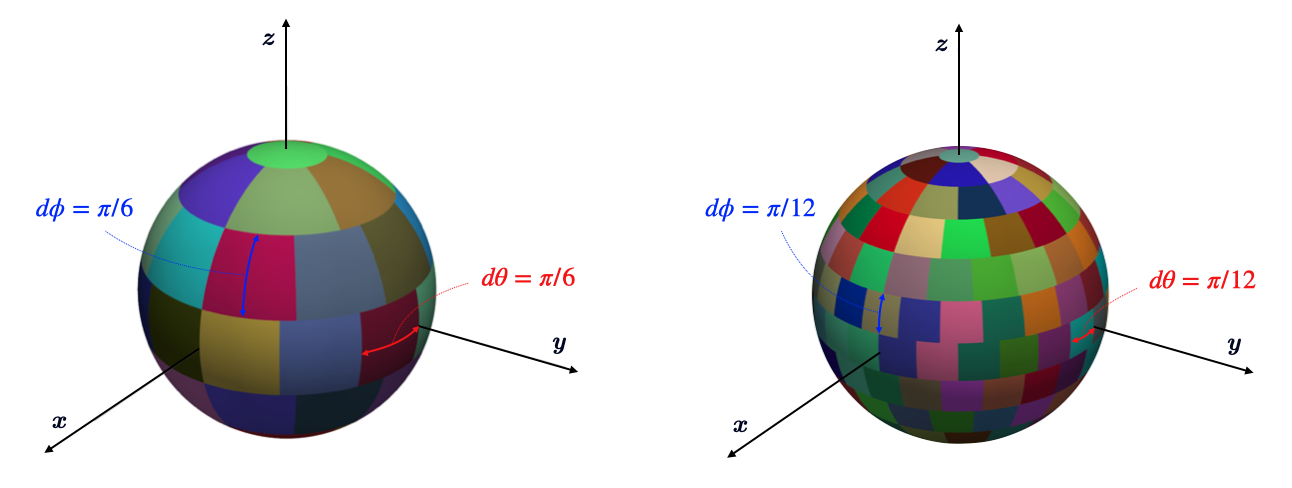}
    \caption{Illustration of the angular partitioning in EPG for different values of $d\theta$ and $d\phi$.}
    \label{fig:grid}
\end{figure}

\begin{equation}
\label{eq1}
    \left.
    \begin{array}{ll}
    \vspace{1ex}
    l = \floor*{\frac{\phi}{d\phi}}    \: , \\
    \vspace{1ex}
    d\theta_l = d\theta * \mathrm{cos}\left((l + 0.5) * d\phi\right)    \: , \\
    m = \floor*{\frac{\theta}{d\theta_l}}    \: .
    \end{array}
    \right.
\end{equation}

We end up with five integer keys $\left(i, j, k, l, m\right)$ that are used in a hash map to save the pose and embeddings. In our experiments, we use the different values for indoor and outdoor datasets to tailor different needs. On ScanNet, $dl=0.4$, and on KITTI $dl=2.0$, while $d\theta = \pi / 6$ and $d\phi = \pi / 6$ on both datasets.

\begin{figure}[t]
    \centering
    \adjincludegraphics[width=0.99\columnwidth,trim={{.001\width} {.12\height} {.48\width} {.001\height}},clip]{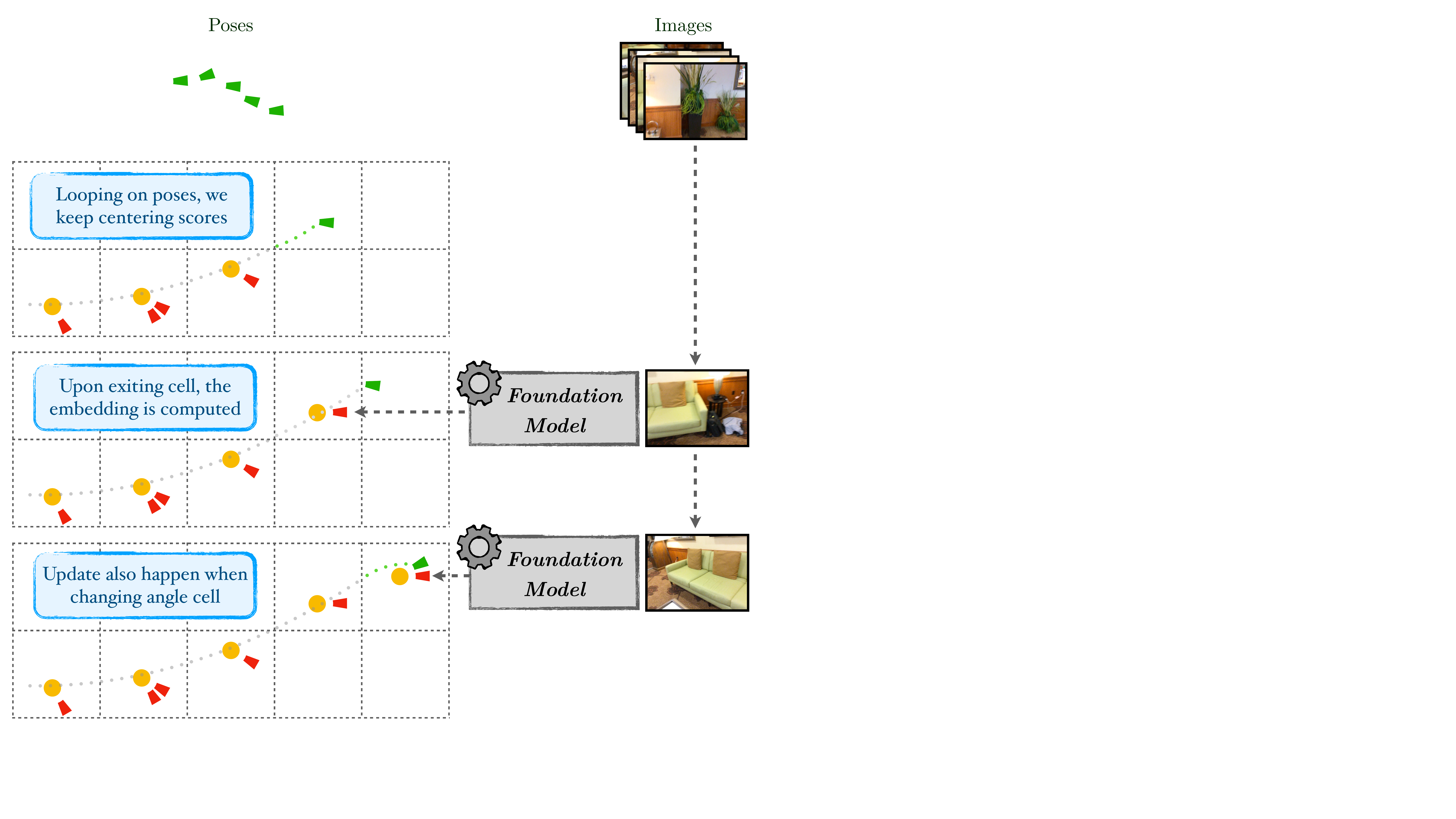}
    \caption{Illustration of the building process of an EPG.}
    \label{fig:build}
    \vspace{-3ex}
\end{figure}

\subsection{Efficient Construction Process}

Given the potential redundancy in the captured image data and considering the computational demands of foundational models, our construction algorithm prioritizes efficiency through selective feature computation. Rather than processing every image sequentially, we compute and update a cell's embedding only when the associated pose exits its boundary, using the image nearest to the cell's center. This approach not only reduces the computational load but also aids in maintaining a focused representation of the pose graph \Cref{fig:build}.

To build the pose graph, we iterate over every image and pose. While traversing a cell, we maintain a ``cell centering score'', evaluating the proximity of each pose to the cell's ideal center. Upon exiting a cell, the embedding corresponding to the pose with the highest centering score within that cell is computed and stored. Additionally, to prevent excessive updates due to trajectory noise, we impose a temporal threshold; a cell will not be updated again if revisited within a short interval, set here as 10 seconds.

We chose to use features from two different models in our EPG. These features can be changed at will, depending on the application's needs. For semantic understanding within the EPG, we integrate features from a CLIP model. In all our experiments, we use the OpenCLIP \cite{ilharco_gabriel_2021_5143773} model ViT-L/14 trained on DataComp-1B \cite{gadre2023datacomp}, as it provides a good trade-off between near state-of-the-art performance and efficiency. For each pose, the semantic embedding is thus a vector of 768 float16 values.

For localization within the EPG, we adopt embeddings similar to those used in AnyLoc \cite{keetha2023anyloc}. We use the DinoV2 ViT-g/14 distilled model with registers and extract the ``value'' features from layer 31. We then aggregate these features with VLAD \cite{jegou2010aggregating}. Using VLAD with 32 centers, we obtain a 32768-dimensional descriptor that we compress into 512 values using PCA for a more compact representation. We create dataset-specific VLAD vocabularies using images from our datasets. Similarly, we fit the PCA transform to the data from each domain. In the experimental section, we provide ablation studies comparing domains for the VLAD vocabulary and PCA dimensions.

Upon completing this process, we obtain a sparse yet comprehensive representation of the 3D environment, encapsulated within a few thousand poses, each appended with crucial environmental embeddings. This stands in contrast to traditional dense representations such as meshes or point clouds, which may contain millions of points. The EPG thus presents a more manageable and computationally efficient framework for spatial representation in robotics applications.

\section{EPG, A VERSATILE TOOL}
\label{sec:using}

With the EPG established as a compact and efficient representation of a 3D environment, we explore its applicability to various robotic tasks. The diverse capabilities of the EPG, which include open-vocabulary queries, disambiguation, language-driven navigation, image-based querying, and re-localization, emphasize its potential as a foundational block for enhancing spatial understanding and operational efficiency in complex environments.

\subsection{Open-vocabulary Queries}

The EPG supports open-vocabulary querying, enabling robots to interpret and act on natural language commands. This functionality leverages the semantic embeddings attached to poses within the graph, using cosine similarity between the text embedding of the query and the pose embeddings to identify the best match. The search is carried out by a vector database tool \cite{douze2024faiss} for efficiency in large-scale scenarios. This approach allows a robot to know where it has seen a certain query, and approximately in which direction. Queries can be for specific objects, such as ``Find a remote control.'' or ``Where is my red backpack?'', or they can define abstract notions, such as ``Is there a trip hazard?'' or ``Where can  I find a place to relax?''. Note that the actual query is extracted from the user command with an LLM oriented by basic prompt engineering. Although the result is not a precise 3D localization of the query, it is sufficient for most robotic applications as a global localizer. When approaching this approximate global localization, other methods for short-range perception and interaction are usually necessary.

\subsection{Disambiguation}

Building upon open-vocabulary querying, EPG offers disambiguation capabilities in scenarios where a query may refer to multiple instances within an environment. Despite the lack of explicit object segmentation or recognition, EPG can propose several probable locations by analyzing the field of view (FOV) overlap between poses. If the proposed locations' FOVs do not overlap, suggesting they may reference distinct objects, the system prompts the user for further information to refine the query. This feature significantly enhances the robot's ability to navigate and interact within densely populated or cluttered spaces and provides an intuitive way to improve user-robot interactions.

\subsection{Language-Directed Navigation}

Language-directed navigation extends the concept of open-vocabulary querying by not only identifying a relevant pose but also guiding a robot to that specific location. Thanks to EPG, the endpoints for navigation are guaranteed to be previously visited positions and thus represent valid navigation goals. This eliminates the need for convoluted calculations to determine the final positioning of the robot around a queried object. We can even push the concept further and use EPG poses as waypoints to navigate to the goal, in a setup similar to teach-and-repeat \cite{furgale2010visual}.

\subsection{Image-Based Queries}

EPG also offers the possibility to locate a specific position within the environment by providing an image. This image-based query capability makes use of the robust localization features within EPG to identify a matching pose even with significant viewpoint variation. This type of query is based on the same cosine similarity concept as text queries but uses the localization features exclusively. The most straightforward applications of this are visual place recognition and re-localization, where the robot is rebooted in an unknown position and needs to localize itself again. However, other applications could include visual similarity search, where an image is provided to the robot to search for similar objects in the environment.

\subsection{Re-Localization}

In the context of robot re-localization, EPG provides a fast solution to retrieve the closest pose to the robot with an image-based query. Due to the robustness of foundation model features for localization, this can even produce an accurate result when the EPG's closest pose is far from the actual robot pose. We propose to go further and add a bundle consistency check for an even more robust re-localization. The idea is to let the robot move a little and aggregate $K_b = 15$ successive poses with local odometry. For each pose, we retrieve the $K_c = 5$ best image-query candidates. Among these $K_b \times K_c$ candidates, there are usually multiple good estimates, thanks to the view redundancy in EPG. We thus leverage a spatial voting scheme to find the most represented pose.

\begin{figure}[t]
    \centering
    \adjincludegraphics[width=0.99\columnwidth,trim={{.001\width} {.454\height} {.590\width} {.001\height}},clip]{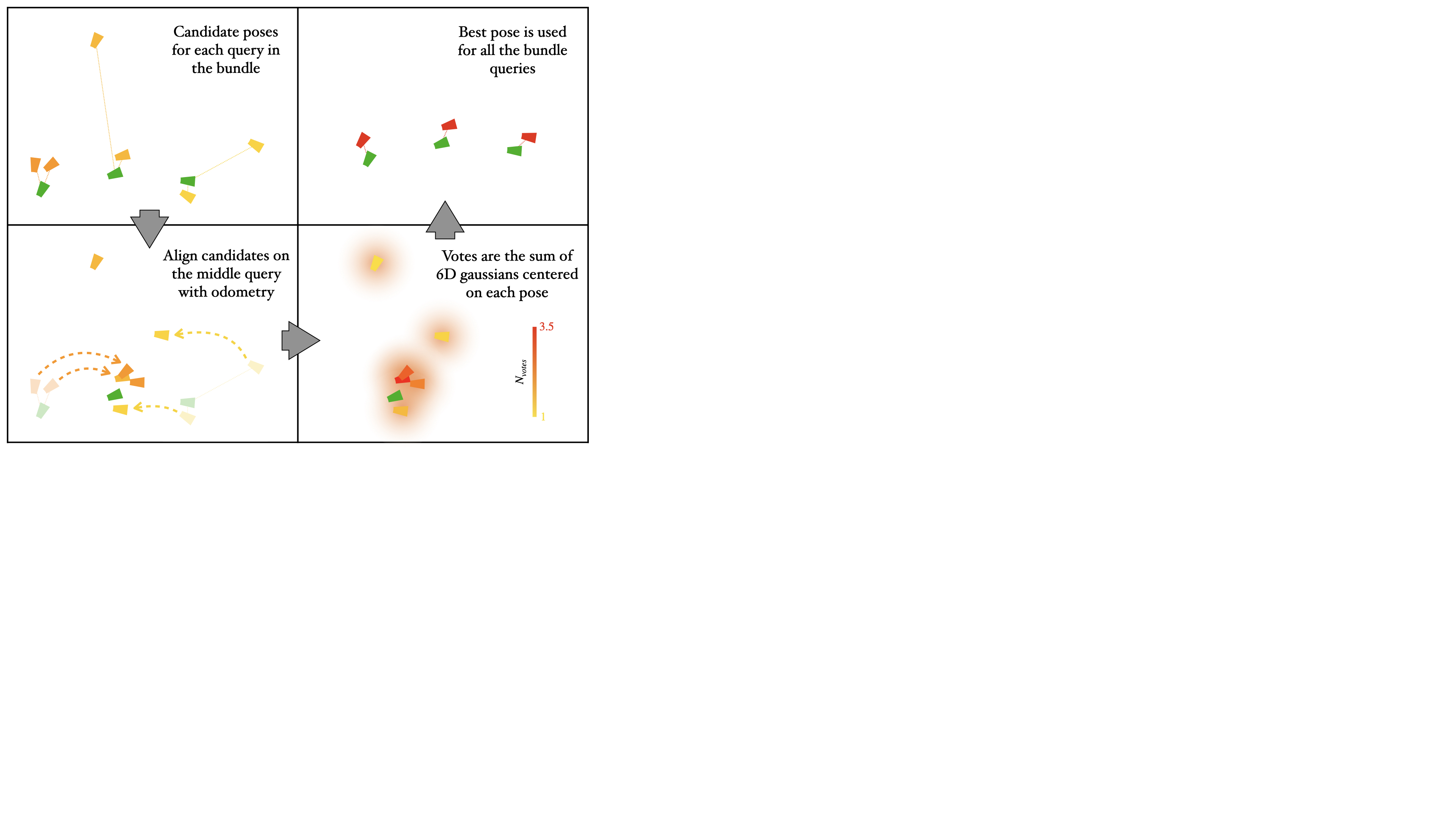}
    \caption{Our Spatial Gaussian Voting scheme for bundle re-localization illustrated with a $K_b=3$ bundle size and $K_c=2$ candidates for clarity. }
    \label{fig:bundle}
    \vspace{-3ex}
\end{figure}

Our Spatial Gaussian Votes are computed following the pipeline shown in \Cref{fig:bundle}. First, we obtain the $K_b \times K_c$ candidates. Then, using the local odometry, we realign the $K_c$ candidates of each query pose to match the middle pose of the bundle. As a result, all the $K_b \times K_c$ pose candidates now represent votes for the best localization for the middle pose of the bundle. We aggregate the votes as a sum of Gaussians in the 5D space (defined by the parameters $\sigma_{xyz}$ and $\sigma_{\theta\phi}$) and determine the pose with the highest vote score as the final estimate. This method also offers a convenient way to gauge the confidence of the estimation compared to cosine similarity values, which are less informative. The higher the vote score, the more reliable the re-localization will be.
We choose the Gaussian parameters empirically: $\sigma_{xyz} = 0.45$m for ScanNet, $\sigma_{xyz} = 2.2$m for KITTI, and $\sigma_{\theta\phi}=20^{\circ}$ for both.

In addition, when a scene mesh and depth images are available, we propose an improved version of our bundle re-localization method. Instead of solely relying on image queries in EPG for pose estimation, we employ Iterative Closest Point (ICP) for local refinement. Each $K_c$ candidate is considered as an initial global pose estimation, and ICP refines this pose to provide more precise localization. Using these refined pose estimations in our Spatial Gaussian Voting mechanism makes it more robust, especially when the robot is far from the EPG. This multi-stage process illustrates the strength of EPG as both a standalone spatial representation tool and a component in comprehensive localization systems.

In summary, EPG presents a versatile and effective foundation for various robotic applications, offering novel solutions for navigating and understanding complex 3D environments. Through the tasks outlined above, EPG demonstrates its potential to significantly advance the field of robotics, enhancing both the efficiency and intuitiveness of robot operations in diverse settings.

\section{EXPERIMENTS}
\label{sec:exp}


\subsection{Datasets}

\begin{figure*}[b!]
    \vspace{-3ex}
    \centering
    \adjincludegraphics[width=0.99\textwidth,trim={{.001\width} {.001\height} {.001\width} {.001\height}},clip]{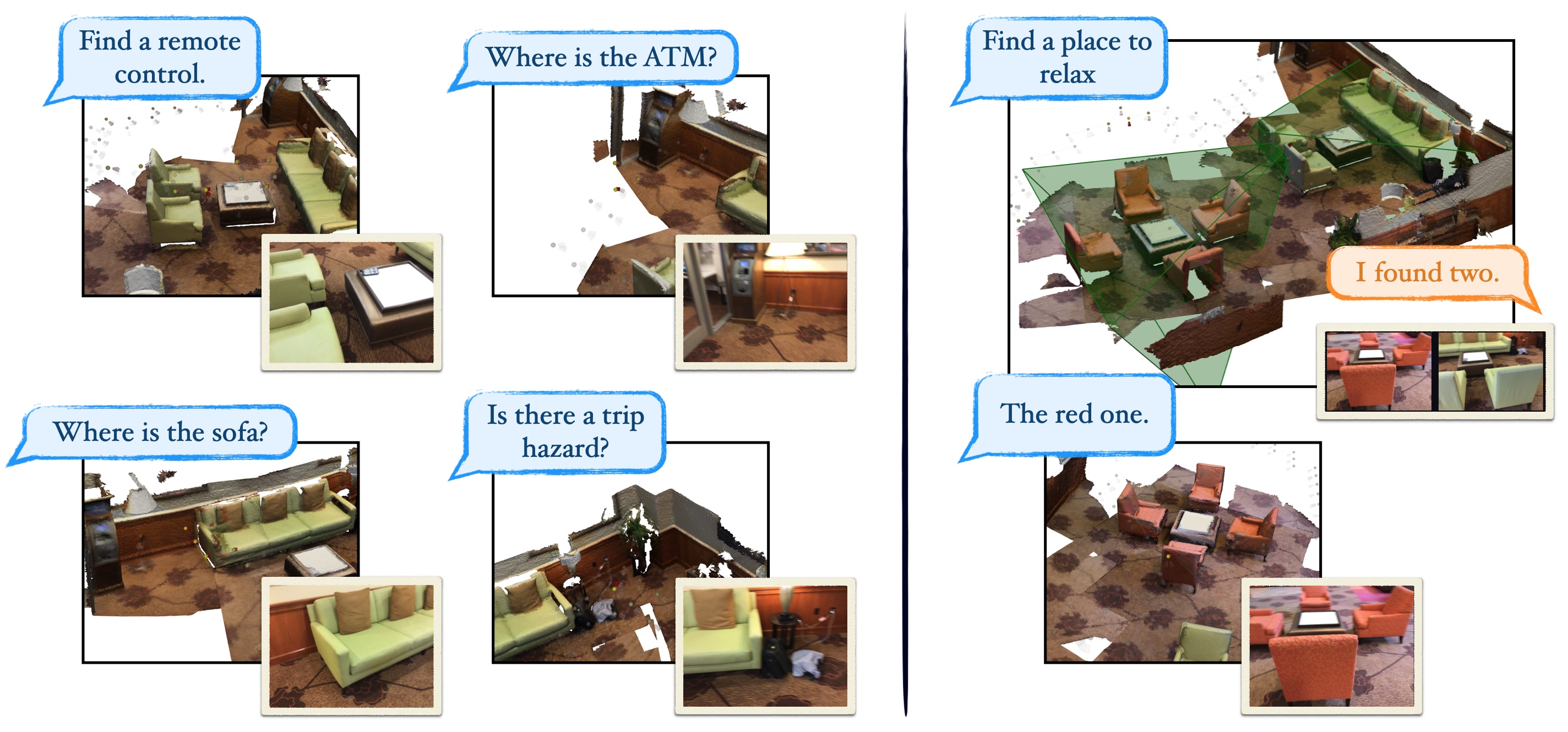}
    \caption{Examples of text queries in an EPG, including concrete objects, abstract notions, and disambiguation interactions.}
    \label{fig:text_queries}
    \vspace{-3ex}
\end{figure*}

In the evaluation of our pose graph estimation and re-localization methods, we use subsets of two widely recognized datasets in the robotics and computer vision communities: ScanNet and KITTI. These datasets provide diverse environments and conditions, which are essential for demonstrating the robustness and versatility of our approach.

\subsubsection{ScanNet}
ScanNet is a richly annotated dataset of 3D scans of indoor scenes \cite{dai2017scannet}. For our experiments, we selected a subset of 16 of the largest scenes that encompass a variety of room types and lighting conditions. This diversity ensures that our methods are tested against common indoor navigation challenges and that our results are broadly applicable to indoor robotics applications. To select the scenes, we compute the bird's-eye view area of each mesh by projecting it onto a binary image and counting the positive pixels. For each scene, starting with the largest one, we align the other scenes representing the same room using ICP initialized from the four 90-degree rotations around the $z$-axis. We select the best alignment manually and repeat the process until we have the following 16 pairs of aligned scenes:

\begin{multicols}{2}
\begin{enumerate}[1.]
\item (0588\_00, 0588\_01)
\item (0667\_00, 0667\_01)
\item (0152\_02, 0152\_01)
\item (0592\_00, 0592\_01)
\item (0678\_00, 0678\_02)
\item (0500\_00, 0500\_01)
\item (0411\_01, 0411\_00)
\item (0520\_01, 0520\_00)
\end{enumerate}
\columnbreak
\begin{enumerate}[1.]  \setcounter{enumi}{8}
\item (0312\_01, 0312\_02)
\item (0334\_02, 0334\_00)
\item (0000\_02, 0000\_01)
\item (0151\_00, 0151\_01)
\item (0665\_00, 0665\_01)
\item (0673\_05, 0673\_00)
\item (0387\_01, 0387\_02)
\item (0038\_00, 0038\_01)
\end{enumerate}
\end{multicols}

\subsubsection{KITTI}
The KITTI dataset is a collection of various sensor data from a vehicle navigating through urban environments \cite{geiger2012we}. To evaluate our re-localization methods, we have divided the dataset into six pairs of sequences:

\begin{enumerate}[1.]
\item (seq00$\left[0, 1233\right]$, seq00$\left[1334, 4540\right]$)
\item (seq02$\left[0, 4030\right]$, seq02$\left[4131, 4660\right]$)
\item (seq05$\left[0, 1131\right]$, seq05$\left[1232, 2760\right]$)
\item (seq06$\left[0, 676\right]$, seq06$\left[777, 1100\right]$)
\item (seq00$\left[0, 1233\right]$, seq07$\left[0, 1101\right]$)
\item (seq09$\left[0, 1591\right]$, seq10$\left[0, 1201\right]$)
\end{enumerate}

For the first four pairs, we split sequences with loop closures into two distinct but overlapping parts. For the last two pairs, we use the available global GPS coordinates to determine the overlap between sequences.

We plan to release both our ScanNet and KITTI re-localization benchmarks alongside this paper.

\subsection{EPG size and view redundancy}

We investigate the construction mechanism of EPGs and provide insights into the structure of this representation. For each dataset, we build a common large-scale EPG using all combined scenes. For ScanNet, since all scenes are centered, we introduce offsets in the $x$ and $y$ directions to distribute the scenes within a shared space and ensure that two distinct scenes do not overlap. For KITTI, we use the global GPS coordinates. The data used to compute the VLAD vocabulary and PCA reduction is specific to each dataset. For both datasets, we leverage the images utilized to build the EPGs.

The EPG is a compact representation that can be tuned to accommodate various application needs. By employing larger subsampling parameters, we can decrease the number of stored poses and thus the redundancy of the associated viewpoints. In \Cref{table:redun}, we demonstrate this concept using a custom metric to quantify redundancy. For both datasets, we use the scene point clouds, which are derived from the mesh in ScanNet and by merging LiDAR point clouds in KITTI. Utilizing the camera's extrinsic and intrinsic parameters, we determine the set of 3D points visible in each view. The overlap value $\mathcal{O}(v_1,v_2)$ between two views $v_1$ and $v_2$ is calculated as the intersection over the union of the two sets of visible points. For an entire EPG, we define the redundancy index $\mathcal{R}_o$ as the average number of overlapping views:

\begin{equation} \label{eq:5}
    \mathcal{R}_o = \frac{1}{N_p} \sum_{v_1} \sum_{v_2 \neq v_1} \left(\mathcal{O}(v_1,v_2) > \frac{o}{100} \right)       \: ,
\end{equation}

\noindent
where $N_p$ represents the total number of poses in the EPG. In the following, we calculate $\mathcal{R}_{50}$ and $\mathcal{R}_{25}$ across all scenes and sequences collectively for each dataset. As shown in \Cref{table:redun}, the redundancy index can be considerable, with more than 10 overlapping views on average within the EPG, even while maintaining reasonable subsampling parameters. Outdoor scenarios are less redundant compared to indoor scenarios because the views are captured from a vehicle traveling on straight roads.

\begin{table}[t]
\caption{EPG statistics for different subsampling parameters on our full ScanNet and KITTI re-localization datasets.}
\begin{center}

\begin{tabular}{ c C{1.9cm}  C{0.8cm} C{1.0cm} C{0.8cm} C{0.8cm}}

 & $\left(dl, d\theta , d\phi\right)$ & $N_p$ &  Size	 & $\mathcal{R}_{50}$ & $\mathcal{R}_{25}$	\TBstrut\\
 
\hline

\multirow{4}{*}{\rot{ScanNet}}  & $\left(0.2 , \pi/6 , \pi/6\right)$ & 4732 & $11.6$MB & 5.80 & 18.65 \Tstrut\\
 & $\left(0.4 , \pi/6 , \pi/6\right)$ & 2610 & $6.4$MB & 2.92 & 9.70 \\
 & $\left(0.8 , \pi/6 , \pi/6\right)$ & 1541 & $3.8$MB & 1.30 & 4.84 \\
 & $\left(0.8 , \pi/3 , \pi/3\right)$ & 1040 & $2.5$MB & 1.15 & 3.87 \Bstrut\\
								
\hline		
					
\multirow{4}{*}{\rot{KITTI$\:$}}  & $\left(1.5 , \pi/6 , \pi/6\right)$ & 6059 & $14.7$MB & 1.42 & 3.49 \Tstrut\\
 & $\left(3.0 , \pi/6 , \pi/6\right)$ & 3493 & $8.5$MB & 0.57 & 1.81 \\
 & $\left(6.0 , \pi/6 , \pi/6\right)$ & 1911 & $4.7$MB & 0.31 & 0.69 \\
 & $\left(6.0 , \pi/3 , \pi/3\right)$ & 1873 & $4.6$MB & 0.29 & 0.69 \Bstrut\\
 				
\hline

\end{tabular}
\end{center}
\label{table:redun}
\vspace{-3ex}
\end{table}


\subsection{Text-related tasks}

We present qualitative examples of text queries and disambiguation to demonstrate the capability of EPG to interact with a user. In \Cref{fig:text_queries}, we depict the outcomes of various text queries, encompassing both concrete objects and abstract concepts. Additionally, we provide an instance of a disambiguation interaction. In our real setup, we allow the user to chat with an LLM agent that extracts the queried object or notion from the user message with simple prompt tuning. We showcase this chat interaction, with text query and disambiguation in the supplementary video.



\subsection{Re-localization}

In this experiment, we evaluate the re-localization performance of EPG on our two datasets. As previously mentioned, we utilize a single large-scale EPG for each dataset, constructed from all combined scenes. This approach is more challenging than testing individual scenes separately because it requires the re-localization method to identify the correct pose across multiple rooms or environments. We employ Recall @ K \cite{zaffar2021vpr, keetha2023anyloc} as the evaluation metric, where a higher recall score signifies better performance. To determine an accurate match, we consider the Cartesian distance between poses $\Delta xyz$ and the angular difference between poses $\Delta\alpha$. We establish two precision levels (fine and coarse re-localization) and adjust the thresholds according to the environment's scale. For the indoor ScanNet dataset, we set the coarse threshold at $\Delta xyz = 0.8$m and the fine threshold at $\Delta xyz = 0.3$m. For the outdoor KITTI dataset, we use larger values with a coarse threshold of $15$m and a fine threshold of $3$m.



 

 




\subsubsection{ScanNet}

In our version of the ScanNet dataset, the second scene from each pair is used as a query sequence. To prevent querying poses that are too far away from the EPG, we employ the coarse threshold to filter them. Additionally, we minimize redundancy in the query sequence by selecting poses that are at least $0.3$ meters or $20$ degrees apart. We assess our re-localization method across four distinct setups: simple image query, bundle query, ICP-refined query, and ICP-refined bundle query. These are compared with a state-of-the-art re-localization technique \cite{keetha2023anyloc} for which we tested their open-source implementation on our dataset. The findings are presented in \Cref{table:reloc_scannet}. Initially, we observe that our simple method, despite utilizing fewer features, outperforms AnyLoc. The addition of our bundle voting scheme significantly enhances the results (by $+17.5$ in coarse R@1 and $+29.1$ in fine R@1). Introducing ICP refinement benefits the simple queries in terms of fine R@1 but not for coarse R@1, which can be explained by the fact that coarse alignments do not consistently provide suitable initial guesses for ICP refinement. Applying bundle optimization to ICP-refined queries yields the best results overall, advancing the simple query by $18.5$ points in coarse R@1 and by $52.7$ points in fine R@1. In addition to these quantitative evaluations, we include qualitative examples of re-localization in the supplementary video.

\begin{table}[t]
\caption{Re-localization results on ScanNet and KITTI dataset.}
\begin{center}
\begin{tabular}{ c l  c  c  c c  }

 & & \multicolumn{2}{c}{coarse} & \multicolumn{2}{c}{fine} \Tstrut\\

 & & R@1 & R@5 & R@1 & R@5 \Tstrut\\

\Xhline{2\arrayrulewidth}

\multirow{6}{*}{\rot{Scannet}} & Anyloc \cite{keetha2023anyloc} & $  71.6$  & $  80.7$  & $  26.4$  & $  36.7$  \Tstrut\\

 & Anyloc-PCA \cite{keetha2023anyloc} & $  68.1$  & $  79.5$  & $  24.0$  & $  36.0$  \Bstrut\\

\cline{2-6}



 & EPG simple        & $  73.5$  & $  81.6$  & $  26.3$  & $  37.1$  \Tstrut\\
 & EPG bundle        & $  90.0$  & $  \mathbf{92.4}$  & $  55.4$  & $  67.9$  \\
 & EPG icp           & $  68.1$  & $  79.8$  & $  54.8$  & $  64.4$  \\
 & EPG ICP-bundle    & $  \mathbf{91.0}$  & $  91.9$  & $  \mathbf{79.0}$  & $  \mathbf{81.1}$  \Bstrut\\

\Xhline{2\arrayrulewidth}

\multirow{4}{*}{\rot{KITTI$\:$}} & Anyloc \cite{keetha2023anyloc} & $  75.8$  & $  77.3$  & $  63.0$  & $  66.1$  \Tstrut\\
 & Anyloc-PCA \cite{keetha2023anyloc} & $  75.8$  & $  77.0$  & $  60.5$  & $  65.6$ \Bstrut\\

\cline{2-6}


 & EPG simple        & $  77.0$  & $  79.3$  & $  64.5$  & $  66.1$  \Tstrut\\
 & EPG bundle        & $  \mathbf{89.0}$  & $  \mathbf{89.3}$  & $  \mathbf{82.4}$  & $  \mathbf{83.2}$  \Bstrut\\

\Xhline{2\arrayrulewidth}

\end{tabular}
\end{center}
\label{table:reloc_scannet}
\vspace{-3ex}
\end{table}

\subsubsection{KITTI}

In our KITTI evaluation, we do not assess ICP-refined bundle queries due to the absence of depth images. The EPG is constructed using the first sequence in each pair, ensuring there are no repetitions when the same sequence occurs in multiple pairs. We follow the same data preprocessing protocol as with the ScanNet dataset, filtering out poses that are too distant from the EPG using the coarse threshold and selecting query poses that are either at least $3.0$ meters apart or differ by $10$ degrees or more. The results are presented in the second part of Table \ref{table:reloc_scannet}. We observe trends similar to those in the ScanNet dataset; our simple query strategy surpasses AnyLoc, and our bundle approach significantly improves the results (by $+12.0$ in coarse R@1 and $+17.9$ in fine R@1). 

\begin{figure}[b]
    \vspace{-2ex}
    \centering
    \adjincludegraphics[width=0.99\columnwidth,trim={{.001\width} {.032\height} {.001\width} {.029\height}},clip]{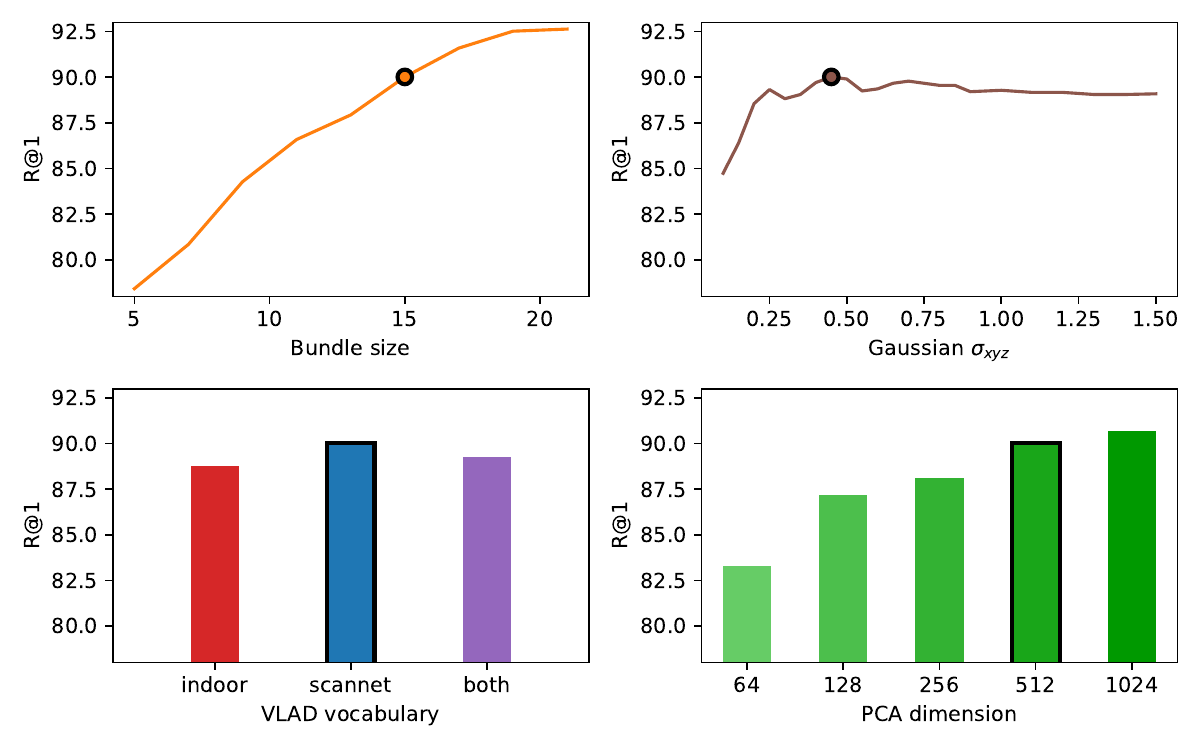}
    \caption{Ablation Study of \textit{EPG Bundle} on Scannet.}
    \label{fig:abl}
\end{figure}

\subsubsection{Ablation studies on Scannet}

In this section, we conduct experiments on the ScanNet dataset using the coarse R@1 metric without performing ICP refinement to obtain more generally applicable results. Similar findings are observed on the KITTI dataset.
We present ablation and parameter studies in \Cref{fig:abl}. Initially, we observe that larger bundles enhance re-localization performance; however, they also result in longer delays before re-localization can occur. We identify 15 as a good trade-off value. Subsequently, we find that a larger Gaussian $\sigma_{xyz}$ facilitates the aggregation of neighboring votes, but excessively increasing this value may disproportionately favor outliers. Additionally, we compare our dataset-specific VLAD vocabularies to domain-specific ones \cite{keetha2023anyloc}. We note better performances for our dataset-specific vocabularies, showing that our 16 ScanNet scenes provide sufficient diversity for a high-quality vocabulary. Data from other datasets can be considered out-of-distribution and degrade the results. Lastly, higher PCA dimensions correlate with improved results but also with increased memory requirements. We settle on 512 as a balanced trade-off for our experiments.

\section{CONCLUSIONS}

In this paper, we introduced the Embedding Pose Graph (EPG), a novel representation that combines foundation model features with a 3D pose graph to create a compact and versatile tool for robotics applications. EPG marks a significant step forward in enabling robots to efficiently understand and navigate large-scale 3D spaces. We demonstrated how EPG supports a variety of tasks such as open-vocabulary querying, disambiguation, image-based querying, language-directed navigation, and re-localization, highlighting its potential to fundamentally change how robots interact with complex environments.

Looking forward, we identify several promising directions for further enhancing EPG's capabilities. Integrating EPG with SLAM pose graph optimization processes could enable dynamic real-time updates of the environment, facilitating lifelong robot applications. Furthermore, combining EPG with large language models (LLMs) would unlock new levels of spatial understanding in a multimodal setup, allowing robots to interact more naturally and effectively with users and their surroundings. Expanding EPG's applications to include tasks such as visual question answering (VQA) could also broaden the scope of human-robot interaction, making robots increasingly useful in practical settings. Overall, EPG establishes a foundation for the development of more intelligent and capable robotic systems, and we look forward to its future advancements.



\bibliographystyle{IEEEtran}
\bibliography{IEEEabrv, main}

\end{document}